# Tac3D: A Novel Vision-based Tactile Sensor for Measuring Forces Distribution and Estimating Friction Coefficient Distribution*

Lunwei Zhang, Yue Wang, Yao Jiang

*Abstract*— The importance of force perception in interacting with the environment was proven years ago. However, it is still a challenge to measure the contact force distribution accurately in real-time. In order to break through this predicament, we propose a new vision-based tactile sensor, the Tac3D sensor, for measuring the three-dimensional contact surface shape and contact force distribution. In this work, virtual binocular vision is first applied to the tactile sensor, which allows the Tac3D sensor to measure the three-dimensional tactile information in a simple and efficient way and has the advantages of simple structure, low computational costs, and inexpensive. Then, we used contact surface shape and force distribution to estimate the friction coefficient distribution in contact region. Further, combined with the global position of the tactile sensor, the 3D model of the object with friction coefficient distribution is reconstructed. These reconstruction experiments not only demonstrate the excellent performance of the Tac3D sensor but also imply the possibility to optimize the action planning in grasping based on the friction coefficient distribution of the object.

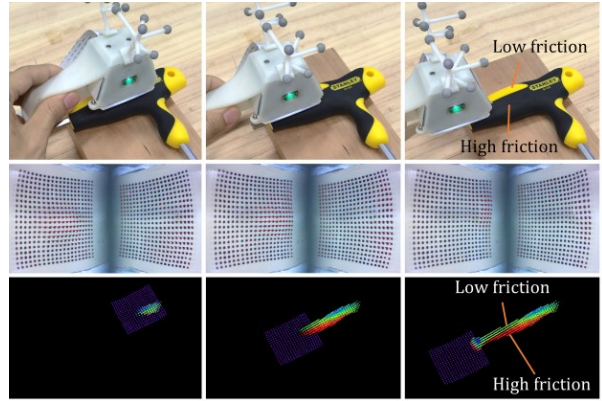

Figure 1. The friction coefficient distribution is estimated with the surface shape and the contact force distribution measured by the Tac3D sensor.

## I. INTRODUCTION

Tactile perception of the contact forces plays an essential role in dexterous manipulation and exploration for both humans and robots[1]. While grasping an object like a fragile cup, instant force feedback from the sense of touch helps grasp stably and securely without dropping or crushing it[2]. Tactile sensors provide robots with essential tactile information and vastly enhance their manipulation ability in unknown environments[1]. Therefore, the integration and application of tactile sensors on intelligent robots to perceive tactile information, such as surface shape and contact force, have long been research focuses. Since being put forward, the tactile sensors are expected to be precise, fast-response, compact, universal, and low-cost[3]. However, despite the rapid development in recent years, meeting the above requirements remains a challenging task.

This work proposes a novel vision-based tactile sensor, the Tac3D sensor, for easily perceiving various contact information, especially the shape, deformation, and force distribution in the contact region, to fulfill those requirements better. Compared with the existing vision-based tactile sensor[4-7], the most important innovation of the Tac3D sensor is the application of virtual binocular vision[8, 9]. This improvement significantly reduces the difficulty and complexity in reconstructing the 3D displacement field of the fingertip, which is required for calculating the 3D contact force distribution.

With the prototype of the Tac3D sensor, we reconstructed several common objects' shapes with the friction coefficient distribution in further experiments. By sliding the sensor's fingertip across the surface of the objects, friction coefficients in different parts of the contact region are estimated according to the measured surface shape and contact force distribution. The results demonstrate that the comprehensive performance of the Tac3D sensor has reached the level of practical applications. More importantly, this work makes it possible for the robot to plan better motion in manipulation according to the objects' friction coefficient distribution and provides a new way for the robot to explore and perceive the unknown environment.

## II. RELATED WORK

### A. Vision-based tactile sensor

The tactile sensors have received extensive attention from scholars in the past decades for their importance to intelligent robots working in an unstructured environment. Among the existing tactile sensors, the vision-based types attract more attention for their advantages of simple structure, high resolution, and expansibility. A typical vision-based tactile sensor consists of one or more cameras and elastic tactile skin.

*This work was supported by National Natural Science Foundation of China (Grant No. 51705274), and a grant from the Institute for Guo Qiang. Tsinghua University.

L. W. Zhang is with the Institute of Manufacturing Engineering, Department of Mechanical Engineering, Tsinghua University, Beijing, China. (e-mail: zlw21@mails.tsinghua.edu.cn).

Y. Wang is with the Institute of Manufacturing Engineering, Department of Mechanical Engineering, Tsinghua University, Beijing, China. (e-mail: wang-yue19@mails.tsinghua.edu.cn).

Y. Jiang is with the Institute of Manufacturing Engineering, Department of Mechanical Engineering, Tsinghua University, Beijing, China. (Corresponding author to provide e-mail: jiangyaonju@126.com).

When the tactile skin is contacting with an object, the images of the deformed tactile skin are captured to extract the desired tactile information[4].

Most vision-based tactile sensors, such as GelForce[10, 11], GelSight[12-14], GelSlim[15, 16], FingerVision[17, 18], use 2D image collected by a monocular camera to reconstruct tactile information. Usually, the displacement field of the elastic tactile skin is reconstructed firstly according to the image. Then the contact force is solved from the displacement field according to the mechanical model. However, the key problem for those sensors is how to reconstruct the 3D displacement field from 2D image. In the previous researches, various methods were proposed to solve the problem. GelForce used double-layered markers and optimization methods to reduce the normal force noise caused by missing normal deformation information[10]. GelSight introduced an improved stereo-photometric method to achieve 3D shape reconstruction in high resolution[19]. Yuji Ito et al. designed a robotic fingertip filled with colored liquid to obtain depth information with limited accuracy by analyzing the light absorption rate[20]. Besides, Feng Gu et al. used a fixed-focus camera and calculated the normal position of the markers according to the size of the circle of confusion[21, 22]. This type of sensor has advantages of simple structure and small size and, hence, it is easy to be integrated into a multi-finger manipulator. However, they have made certain sacrifices in the complexity of the algorithm and/or the measurement accuracy.

Another type of tactile sensor uses a depth camera instead of a monocular camera to directly obtain the 3D shape and deformation of the tactile skin. Tao Zhang et al. installed two cameras in the sensor and tracked the 3D movement of the markers with binocular vision[23]. Isabella Huang et al. applied a TOF (Time of Flight) camera to directly capture the 3D shape of the inner surface of the fingertip[24]. The application of depth cameras greatly improved the accuracy and robustness in measuring the shape and displacement field and reduced the computational costs of 3D reconstruction. However, these sensors are difficult to be integrated into the manipulator due to the large size of the depth cameras. Besides, the expensive hardware is not conducive for general use.

The intention of designing the Tac3D sensor is to seek to integrate the advantages of the above two types of tactile sensors. The method of virtual binocular vision allows the monocular camera to achieve the exact 3D measurement as that of the binocular camera. It brings the Tac3D sensor the advantages of high accuracy, robustness, and low computational costs while maintaining the characteristics of simple structure, acceptable overall size, and inexpensive.

### B. Friction Coefficient Measurement

With the force distribution provided by the tactile sensor, important information that involves object properties and contact states such as surface texture, friction coefficient, slip, etc., can be further obtained. Knowing these properties and their distribution in grasping helps plan better motion and grip force[2]. Studies have shown that the friction coefficient plays a vital role in grasping and dexterous operations, but measuring the friction coefficient is a challenging task for tactile sensors[1]. Not only the 3D contact force but also surface shapes and normals are required so that tangential and normal forces can be decoupled to estimate the friction coefficient. Only a few of the existing tactile sensors[5, 25] have managed to estimate the friction coefficient at a single contact point, and there is no vision-based tactile sensor. The Tac3D sensor is expected to break through the bottleneck. Due to the measurement capability of the 3D deformation and the contact force distribution, it has been possible to reconstruct the object's shape with the friction coefficient distribution. This information of friction coefficient distribution will help the robot achieve better grasp in determining grasping positions and planning motion.

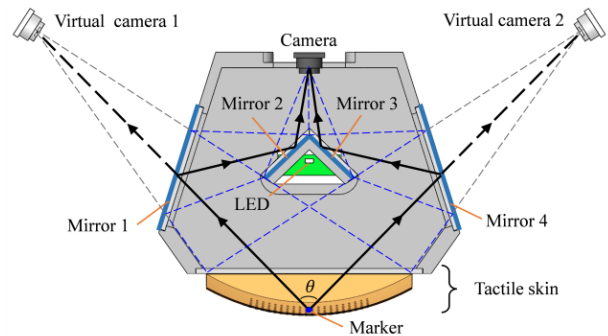

Figure 2. The cross-sectional view and the light path of the Tac3D sensor

### III. SENSOR DESIGN

In this section, the structure of the Tac3D sensor will be introduced in detail, and then the selection of key structural parameters will be discussed.

### A. Structure

The Tac3D sensor tracks three-dimensional markers embedded in transparent silicone by a virtual binocular camera. As shown in Fig. 2, the virtual binocular camera consists of an ordinary camera and several mirrors. The camera captures the markers' virtual images in the mirrors, instead of the actual markers, through two different light paths to simultaneously capture the markers from two different perspectives. From the perspective of the actual camera, the field of view is symmetrically divided into left and right parts, respectively showing the images observed from the positions of the two virtual cameras (Fig. 2). The light paths on both sides are symmetrical. In the left part, light from the markers is reflected by mirrors 1 and 2 before entering the camera, while light on the right is reflected by mirrors 4 and 3. According to the plane mirror imaging principle, it is equivalent to the camera observing the markers at virtual cameras 1 and 2, respectively.

The virtual cameras 1 and 2 form a binocular camera, and the parallax angle is denoted by $\theta$. The positions of the virtual cameras can be calculated from the known positions of the actual camera and the mirrors according to the principle of plane mirror imaging. In this way, the markers' 3D positions and displacements can be measured based on the principle of binocular vision.

The elastic fingertip is another crucial part of the Tac3D sensor (Fig. 3). It includes four layers from the inside to the outside:

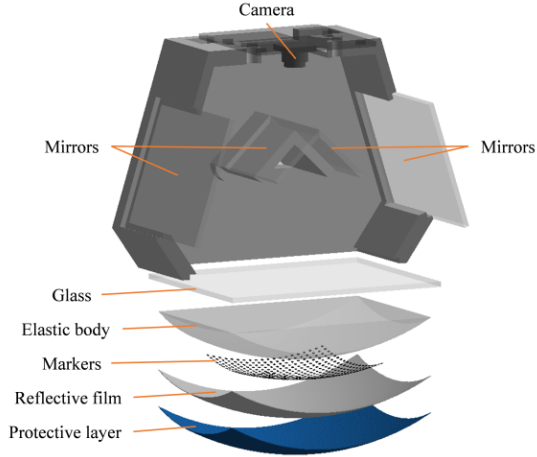

Figure 3. Exploded view of the Tac3D sensor

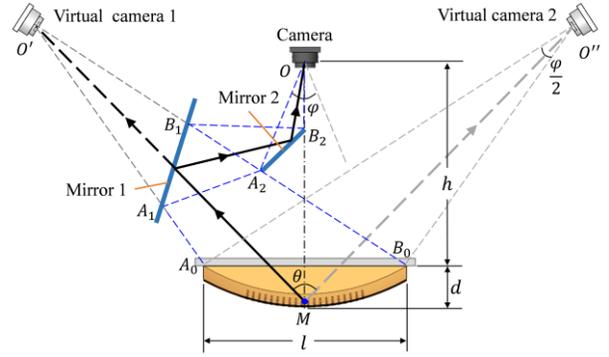

Figure 4. The key structural parameters of the light path. Since the light paths on both sides are symmetrical, the left half of the light path is shown as an example.

1) The **elastic body** made of transparent silicone (Kafuter K705) deforms when forces are applied.

2) The **markers** painted with carbon black move with the deformation of the elastic body.

3) The **reflective film** made of silicone mixed with silver powder isolates the light inside and outside the sensor.

4) The **protective layer** made of polyester textile protects internal structure of the fingertips and prolong the life of the sensor.

A markers array of 20 × 20 is arranged in the markers layer, and the distance between adjacent markers is 1.27 mm. These markers provide a sensing area of approximately 25 mm × 25 mm for the Tac3D sensor. The fingertip has a spherical surface with a radius of about 45mm to make good contact with objects with a large flat surface or even a concave surface.

In particular, the reflective film and the protective layer should be made as thin as possible to ensure the force measurement accuracy. According to the theory of elasticity, the farther the markers are from the contact surface, the more ill-conditioned the mechanical model between the markers' 3D displacements and the contact forces is. Therefore, when the 3D displacements of the markers can be measured, the markers should be as close as possible to the fingertip's surface to improve the robustness of solving the distribution force.

### B. Key structural parameters

The installation positions of the mirrors determine the measurement accuracy of the virtual binocular camera and the compactness of the sensor. Here, the left half of the light path shown in Fig. 4 is taken as an example to illustrate the constraints and key structural parameters for determining the installation position of the mirrors.

There are the following constraints in the symmetric light path:

1) The lights from the two endpoints of the fingertip $A_0$ and $A_1$ enter the camera along the light path $A_0 \rightarrow A_1 \rightarrow A_2 \rightarrow O$ and $B_0 \rightarrow B_1 \rightarrow B_2 \rightarrow O$, respectively. In order to make full use of the camera's field of view, the angle between the two beams of light entering the camera should be half of the camera's field of view angle, denoted by $\varphi$.

2) According to the principle of mirror reflection, when light is reflected at $A_1$, $A_2$, $B_1$ and $B_2$, the exit angles should be equal to the incident angles.

3) In order to make the sensor compact, the mirror 2 should be as short as possible. Therefore, its two endpoints are located exactly at $A_2$ and $B_2$. Point $A_2$ locates on the line $B_0 B_1$ without blocking the light. Point $B_2$ locates on the axis of symmetry $OM$.

4) According to the principle of binocular vision, the parallax angle should be as close as possible to 90° to improve the accuracy of 3D measurement. So here we require that the angle between the connecting lines of two virtual binocular cameras and the central mark point M, namely $\angle O' MO''$, is 90°.

Under the above constraints, the light path can be determined by specifying the distance from the camera to the elastic body, denoted by $h$, the camera's field of view angle, denoted by $\varphi$, the length, denoted by $l$, and thickness, denoted by $d$, of the fingertip. Although it is difficult to solve the analytical solution of the mirrors' position, it is possible to solve a numerical solution with the help of CAD software automatically. Here we specify $l = 40$ mm, $h = 40$ mm, and $d = 8$ mm. In addition, the IMX219-77 camera is selected, of which $\varphi$ is 48°.

## IV. METHOD

### A. Measuring the Distributions of Contact Forces

Firstly, the virtual cameras' parameters should be obtained. The approximate positions of the two virtual cameras can be calculated by the ideal installation positions of the camera and the mirrors. However, because of the distortion and the inevitable installation error of the camera and mirrors, it is necessary to calibrate the virtual camera's external, internal, and distortion parameters before installing the elastic body.

Then, the displacement field of the elastic body, which is required as a boundary condition to calculate the contact force, is reconstructed by tracking the markers. When the elastic body is deformed by force, the markers move synchronously, so their positions and displacements directly indicate the shape and deformation of the elastic body. The position of each marker in the sensor coordinate system can be solved according to the principle of binocular vision. Note that the refraction at the interface of the glass and air results in a deviation in the measured marker's position. The deviation can be calculated numerically, but the process is complicated. Therefore, we used a polynomial to fit the relationship between the deviation and the observed markers' position to quickly compensate for the deviation.

Here, we denote the measured displacement of the markers as a column vector $\boldsymbol{D} = (d_x^1, d_x^2, ..., d_x^N, d_y^1, d_y^2, ..., d_y^N, d_z^1, d_z^2, ..., d_z^N)^T$. $N$ represents the number of the markers. $d_x^i$, $d_y^i$, and $d_z^i$ are the components of the displacement of marker $i$ in the $x$, $y$, and $z$ directions. Assuming that the applied force is uniform in each small area near a marker, similarly, we denote the forces applied on all small areas as a column vector $\boldsymbol{F} = (f_x^1, f_x^2, ..., f_x^N, f_y^1, f_y^2, ..., f_y^N, f_z^1, f_z^2, ..., f_z^N)^T$. The relationship between force and displacement is as the following equation:

$$\boldsymbol{D} = \boldsymbol{H}\,\boldsymbol{F}. \qquad (1)$$

In equation (1), $\boldsymbol{H}$ is a 3N × 3N conversion matrix obtained according to the work of Katsunari Sato et al. [26] but with the finite element method. In this work, the elastic body is discretized into 8-node hexahedron elements with an edge length of one-third of the distance between the markers. Then, the contact force can be calculated by equation (2).

$$\boldsymbol{F} = \boldsymbol{H}^{-1}\,\boldsymbol{D}. \qquad (2)$$

Equation (2) is effective when there are few markers. However, as the number of markers increases, the calculated force becomes sensitive to the measurement noise in the displacement. In order to reduce the influence of noise, Tikhonov Regularization is applied to equation (2) to improve the robustness. The simplified solution expression is shown as equation (3).

$$\boldsymbol{F} = (\boldsymbol{H}^T \boldsymbol{H} + w\,\boldsymbol{I})^{-1}\,\boldsymbol{H}^T\,\boldsymbol{D}. \qquad (3)$$

Here $\boldsymbol{I}$ is an identity matrix of the same dimension as H, and w is a coefficient that should be selected according to the magnitude of the noise in the measured displacement. In equation (3), $(\boldsymbol{H}^T \boldsymbol{H} + w\,\boldsymbol{I})^{-1}\,\boldsymbol{H}^T$ can be calculated in advance and is denoted as $\boldsymbol{K}$. So, equation (3) is simplified to the same form as equation (2), which is shown as equation (4).

$$\boldsymbol{F} = \boldsymbol{K}\,\boldsymbol{D}. \qquad (4)$$

### B. Reconstruction of Friction Coefficient Distribution

With the ability to accurately measure the position of the marker point and the contact distribution force in real-time, the Tac3D sensor provides vital information for estimating the friction coefficient. In the experiment, the Tac3D sensor touches the object fixed on the table, and tactile information is produced and recorded by frame.

The first step is to detect the contact region. We transformed the position data of all the markers into a 3D point cloud, so that the surface normals of the fingertip can be obtained easily with PCL (Point Cloud Library, https://pointclouds.org/). Then the normal and tangential force is decoupled according to the direction of the surface normals. The region where the normal force is greater than a certain threshold is determined as a contact region.

The second step is to collect the surface point cloud. Considering that the reflective film and protective layer are thin, we approximate that the markers in the contact region coincide with the object's surface. In other words, these markers can be regarded as a local point cloud of the object surface. While the tactile sensor sliding across the object, local point clouds are continuously collected, converted to the global coordinate system according to the global position of the tactile sensor, and then combined to form a complete surface point cloud. Theoretically, the required global position of the tactile sensor can be converted from the known end position of the robot arm. In this work, we focus on tactile perception and reconstruction rather than robot control methods. So, we use the OptiTrack motion capture system for global positioning of the Tac3D sensor.

The third step is to resample the point cloud and reconstruct the surface. Through resampling, the point cloud becomes smooth and uniform. Then the Poisson surface reconstruction is applied to transform the point cloud into the triangle mesh.

The final step is to estimate the friction coefficient at each slip point. According to the movement distance of the markers between two adjacent frames in the global coordinate system, the points in contact region that have relative slippage with the surface of the object are filter out. Here we call them slip points, and record the ratio of the tangential force to the normal force as the preliminary estimate of friction coefficient at each slip point. For each surface point, find all the $n$ slip points in the neighborhood of radius r. Then the best estimate of the friction coefficient at point P is calculated by equation (5).

$$\mu = \frac{\sum w_i \mu_i}{\sum w_i}. \qquad (5)$$

In equation (5), $\mu_i$ is the preliminary estimate of friction coefficient at the slip point $i$ in the neighborhood, and $w_i$ is the corresponding weight, calculated by equation (6).

$$w_i = (r - r_i)^\alpha \cdot \mu_i^\beta. \qquad (6)$$

And in equation (6), $r_i$ is the distance from the slip point $i$ to the surface point, $\alpha$ and $\beta$ are adjustable parameters, which should not be negative. $\alpha$ controls the influence of distance on weights. A large $\alpha$ means more consideration is given to the slip points close to the point $P$, while a small $\alpha$ means it is likely to consider all the slip points equally. $\beta$ controls the influence of the preliminary estimate itself on the weight. In order to reduce the influence of a small estimated value when a non-slip point is misjudged as a slip point, a large $\beta$ is usually select. The above method is applied to each surface point to realize the reconstruction of the friction coefficient distribution finally.

## V. RESULT

In this section, verification experiments are carried out based on the prototype. In the first part, we tested the performance, involving real-time performance, the accuracy of the displacement and force measurement, of the Tac3D sensor. In the second part, the experiments of reconstructing the shape and the friction coefficient distribution of various common objects demonstrate the practicability of the Tac3D sensor.

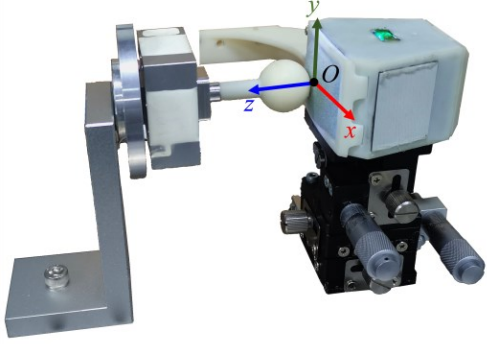

Figure 5. Platform for the validation experiment.

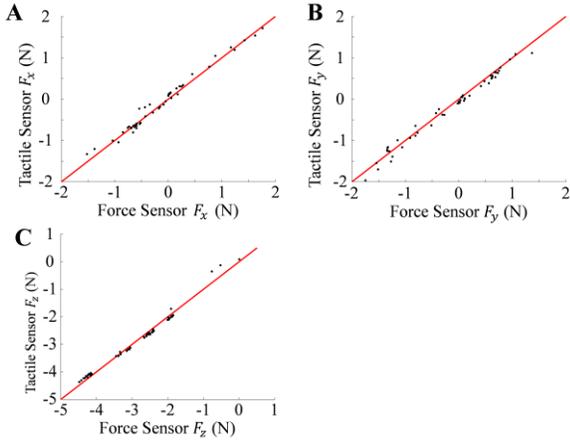

Figure 6. The comparison of the force measured by Tac3D and the ground truth in *x*, *y* and *z* direction. The red lines represent the ideal result, and the black dots represent measured result.

### A. Measurement of Displacement and Force

Firstly, we tested the accuracy performance of the virtual binocular camera. The result is represented by the RMSE (root mean square error) of the measured displacements of the dots on a calibration board. Before installing the soft base components for the Tac3D sensor, we placed a dot calibration board, which is installed on a Micro-motion platform, in the field of view of the virtual binocular camera. During the test, the calibration board is moved 3 mm along *x*, *y* and *z* directions in sequence, and measured displacements of all the dots in each movement are recorded. The RMSE of the measured displacement in each direction is calculated and all less than 0.03 mm.

As the conversion matrix $H$ is obtained through the finite element method, there will inevitably be a deviation from the actual situation. The deviation may result in systematic errors of the measured force. However, it can be corrected by multiplying the $F_x$, $F_y$, and $F_z$ by coefficients $C_x$, $C_y$, and $C_z$, respectively. With a set of various measured resultant forces and the ground truth measurements, we can get $C_x$, $C_y$, and $C_z$ by linear regression. The resultant forces are calibrated and evaluated by summing the components of forces in each direction. The experimental platform shown in Fig. 5 is to calibrate and verify contact force measurement. A three-axis force sensor is fixed on the left side with a sphere on the end to apply a known force. The Tac3D sensor is mounted on the three-axis micro-motion platform on the right side. The micro-motion platform is moved to make the fingertip of the Tac3D sensor and the sphere contact with force in different directions and magnitude. Resultant force is calculated from the measurement of the Tac3D sensor. At the same time, the force sensor provides another measurement, which is regarded as the ground truth within the error permission, to verify the performance of the Tac3D sensor. Fig. 6 shows that the force measured by Tac3D is close to the ground truth after the calibration.

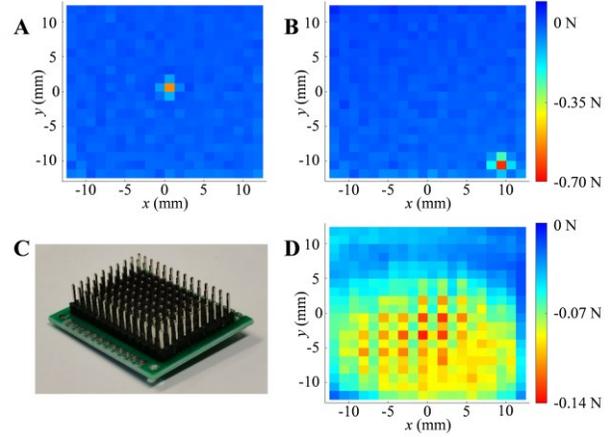

Figure 7. Evaluation of the spatial resolution. (A, B) Constructed normal contact forces when forces are applied through a probe with 2 mm end diameter. (C) The arrayed pins are used to test the spatial resolution of the Tac3D sensor. (D) The reconstructed normal contact forces when forces are applied by the arrayed pins.

As Tac3D is a sensor for measuring contact distribution force, we also need to test its spatial resolution for measuring force distribution. Therefore, we have tested the performance of the Tac3D sensor with single-point contact and multi-point contact experiments. Fig. 7 (A, B) shows the reconstructed z-direction contact force when a probe with a diameter of 2mm is used to apply force to different positions on the sensor in a single-point contact experiment. The sensor accurately measured the location and area of the force applied. Fig. 7(D) shows the reconstructed force distribution when using an array of pins (shown in Fig. 7(C)) to press the sensor. The distance between adjacent pins is 2.54 mm. the result shows that the Tac3D sensor is able distinguish the contact points 2.54mm apart.

The real-time performance of Tac3D tactile sensor is also tested. Considering that it is not cost-effective to equip mobile robots with expensive high-performance computers for tactile perception, we selected the NVIDIA Jetson Nano B10 Developer Kit with a 4-core Cortex-A57 processor core and 4GB LPDDR memory as the computing platform. In order to be compatible with other embedded platforms, GPU is not used in this experiment. The sensor's program is written in C++ with OpenCV (https://opencv.org/), and finally realize the measurement of the force distribution of 24Hz, which can meet the requirement of real-time tactile feedback in grasping control.

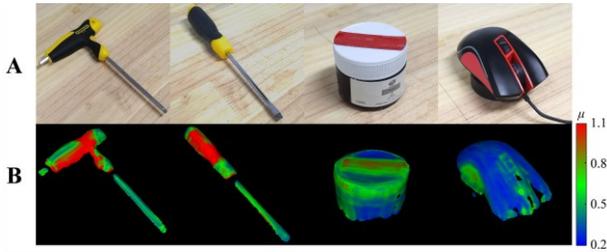

Figure 8. (A) Hexagon wrench, screwdriver, paint can, and mouse used in the reconstruction experiment of surface shape and friction coefficient distribution. (B) The reconstructed surface meshes with friction coefficient distribution.

*B. Reconstruction of Friction Coefficient Distribution*

As shown in Fig. 1, we use the Tac3D sensor to slide along the boundary between two different materials on the surface of a hexagon wrench to observe the perception of friction coefficient in different contact areas through a single touch. The sensor detects a large friction coefficient on half of the surface contacting with rubber (the black part), while a smaller friction coefficient is seen on the half contacting with plastic (the yellow part).

In further experiments, we used the Tac3D tactile sensor to reconstruct common objects' surface shape and friction coefficient distribution, including a hexagonal wrench, a screwdriver, a mouse, a paint can, and a bottle. All the selected experimental objects have surfaces composed of multiple materials and complex shapes. Fig. 8 shows the photos of these items, the point clouds, and the reconstructed surface with friction coefficient distribution.

The parts with large friction coefficients, such as the antiskid rubber on the tools and the Scotch brand tape attached to the top of the can, appear red in the result (the friction coefficient is above 0.8). The parts with a medium friction coefficient, such as the plastic surfaces of the tools and the side of the mouse, appear green (the friction coefficient is between 0.5~0.8). The surfaces with a small friction coefficient, such the sticker on the can and the mouse shell，appear blue (the friction coefficient is 0.5 or less).

As shown in Fig. 9, the Tac3D sensor also has a good reconstruction effect on the bottle's details. The white anti-skid pads on both sides of the bottle cap and the round rubber block with a diameter of about 8mm on the top (shown in yellow outline) appear in the reconstruction result as a red representing a large friction coefficient, which is in sharp contrast with the smooth bottle body around it.

## VI. Conclusion

In this work, we propose a new vision-based tactile sensor, the Tac3D sensor. Compared with the existing tactile sensor, the significant improvement of the Tac3D sensor is the first use of virtual binocular vision in the vision-based tactile sensor to obtain the 3D deformation of the finger. Due to this improvement, the Tac3D sensor achieves accurate force distribution measurement with low computational costs, acceptable overall size, and low cost. Furthermore, based on the Tac3D sensor's ability of the force distribution perception and the surface shape reconstruction, we have realized the measurement and reconstruction of the friction coefficient distribution on the surface of common objects. While proving the excellent performance of the Tac3D sensor, it also provides a new way for robots to explore and perceive the environment.

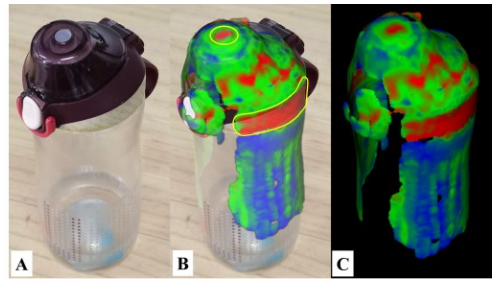

Figure 9. (A) The bottle used in the experiment. (B) The shape and material distribution of the bottle are highly matched with the shape and friction coefficient distribution measured by the Tac3D sensor. (C) The reconstructed bottle surface with friction coefficient distribution.

In the future, we will continue to explore the perception and application of tactile information applications of intelligent robots, especially in the miniaturization of the Tac3D sensor and the robot grasping motion planning based on the friction coefficient distribution.